\documentclass{article}
\usepackage{spconf,amsmath,epsfig}

\usepackage{tikz}

\newcommand\copyrighttext{%
  \footnotesize Copyright 2024 IEEE. Published in the 2024 IEEE International Geoscience and Remote Sensing Symposium (IGARSS 2024), scheduled for 7 - 12 July, 2024 in Athens, Greece. Personal use of this material is permitted. However, permission to reprint/republish this material for advertising or promotional purposes or for creating new collective works for resale or redistribution to servers or lists, or to reuse any copyrighted component of this work in other works, must be obtained from the IEEE. Contact: Manager, Copyrights and Permissions / IEEE Service Center / 445 Hoes Lane / P.O. Box 1331 / Piscataway, NJ 08855-1331, USA. Telephone: + Intl. 908-562-3966.
  DOI: \href{https://doi.org/10.1109/IGARSS53475.2024.10640734}{10.1109/IGARSS53475.2024.10640734}}

\renewcommand\copyrightnotice{%
\begin{tikzpicture}[remember picture,overlay]
\node[anchor=south,yshift=5pt] at (current page.south) {\fbox{\parbox{\dimexpr\textwidth-\fboxsep-\fboxrule\relax}{\copyrighttext}}};
\end{tikzpicture}%
}

\usepackage[baseline]{euflag}
\usepackage{hyperref}
\usepackage{subcaption}
\usepackage{pifont}
\usepackage{algorithm}
\usepackage{algorithmic}
\newcommand{\cmark}{\ding{51}}
\newcommand{\xmark}{\ding{55}}

\title{Evaluating the Efficacy of Cut-and-Paste Data Augmentation in Semantic Segmentation for Satellite Imagery}
\name{Ionut M. Motoi, Leonardo Saraceni, Daniele Nardi, Thomas A. Ciarfuglia
    \thanks{This work received funding by: \\ 
\euflag \quad the European Union’s Horizon 2020 research and innovation programme under grant agreement No 101016906 – Project CANOPIES,\\
\euflag \quad Project AGRITECH Spoke 9 - Codice progetto MUR: AGRITECH ”National Research Centre for Agricultural Technologies” - CUP CN00000022, of the National Recovery and Resilience Plan (PNRR) financed by the European Union ”Next Generation EU”,\\
\hspace*{2em} Sapienza University of Rome as part of the work for project \textit{H\&M: Hyperspectral and Multispectral Fruit Sugar Content Estimation for Robot Harvesting Operations in Difficult Environments}, Del. SA n.36/2022, and project \textit{Weakly and Semi-Supervised Learning for Semantic Segmentation using Satellite Images}, AR123188AA99CE2A.}
    }
\address{Department of Computer, Control and Management Engineering\\ Sapienza University of Rome\\ 00185 Rome, Italy}

\begin{document}
\maketitle
\copyrightnotice
\begin{abstract}
Satellite imagery is crucial for tasks like environmental monitoring and urban planning. Typically, it relies on semantic segmentation or Land Use Land Cover (LULC) classification to categorize each pixel. Despite the advancements brought about by Deep Neural Networks (DNNs), their performance in segmentation tasks is hindered by challenges such as limited availability of labeled data, class imbalance and the inherent variability and complexity of satellite images.
In order to mitigate those issues, our study explores the effectiveness of a Cut-and-Paste augmentation technique for semantic segmentation in satellite images. We adapt this augmentation, which usually requires labeled instances, to the case of semantic segmentation. By leveraging the connected components in the semantic segmentation labels, we extract instances that are then randomly pasted during training.
Using the DynamicEarthNet dataset and a U-Net model for evaluation, we found that this augmentation significantly enhances the mIoU score on the test set from 37.9 to 44.1. This finding highlights the potential of the Cut-and-Paste augmentation to improve the generalization capabilities of semantic segmentation models in satellite imagery.
\end{abstract}

\section{Introduction}
\label{sec:intro}
Semantic segmentation in satellite imagery involves classifying each pixel into categories like impervious surfaces, cultivated areas, forests, or water bodies. This process, often referred to as Land Use Land Cover (LULC) classification, is crucial for interpreting the Earth's surface and finds applications in environmental monitoring and urban planning, among others. However, it faces several challenges, such as the limited availability of labeled data~\cite{song2019survey}, the inherent variability and complexity of satellite imagery, and class imbalances within datasets. Generating pixel-level annotations is not only costly and labor-intensive but also prone to biases, where certain classes are underrepresented, leading to models that struggle with rare categories.

One of the traditional approaches to mitigate these issues has been image augmentation, which generates new training samples through transformations like flips, rotations, crops, and color adjustments, thereby improving model performance and generalization. Instance-level augmentations, such as those based on the Cut-and-Paste concept~\cite{dwibedi2017cut, remez2018learning, fang2019instaboost, ghiasi2021simple}, have shown promise in object detection and instance segmentation tasks in conventional camera imagery, as well as in construction resource detection using UAV-acquired images~\cite{bang2020image}. These methods rely on the idea that diverse object representations can be achieved by manipulating instance placement~\cite{dwibedi2017cut}.

Some techniques further refine this concept by placing object masks in realistic locations to create consistent appearances~\cite{remez2018learning, dvornik2018modeling, fang2019instaboost, dvornik2019importance}, while others paste instances without the need for contextual modeling~\cite{ghiasi2021simple}.

In the context of satellite images, generative adversarial networks have been employed to paste buildings into high-resolution optical images to aid binary change detection models~\cite{chen2021adversarial}. Another approach involves directly pasting buildings onto diverse backgrounds, enhancing building segmentation in very high-resolution images~\cite{illarionova2021object}. Despite these recent advancements, a gap remains in adapting these techniques to the specific challenges of semantic segmentation in satellite imagery. This is mostly because the labels used for this task do not differentiate between individual objects or instances, which are necessary for instance-level augmentations. Moreover, the application of a Cut-and-Paste augmentation technique in medium-resolution satellite images (1-10 meters) has not yet been explored.

To bridge those gaps, we explore the use of a Cut-and-Paste augmentation for semantic segmentation of satellite images. The approach relies on extracting and saving instances from the semantic segmentation labels using connected components analysis. This step allows us to generate augmented images by pasting the extracted instances in various configurations onto the original training images. Unlike other approaches that may restrict the source of instances to the target image itself~\cite{remez2018learning, fang2019instaboost} or to another image randomly selected from the dataset~\cite{chen2021adversarial}, our method can leverage the entire training set, allowing for more diverse compositions. We also allow instances to overlap, creating atypical yet plausible scenarios that encourage models to generalize beyond frequent patterns.

This technique can be categorized as model-free, multi-image, and instance-level, as defined by Xu et al.~\cite{xu2023comprehensive} since it does not rely on additional generative models, uses multiple images to produce augmented samples, and focuses on manipulating specific instances. Our approach offers a simple solution for generating new semantic segmentation data of satellite images without requiring additional manual annotations. Hence, it addresses the previously mentioned challenges by enhancing data diversity, which is crucial for training robust semantic segmentation models.

Our evaluation, using the DynamicEarthNet dataset and a U-Net model baseline, demonstrates that the Cut-and-Paste augmentation significantly improves the mean Intersection over Union (mIoU) score on the test set from 37.9 to 44.1, confirming its practical benefits.

\begin{figure}[t]
    \centering
    \includegraphics[width=0.91\linewidth]{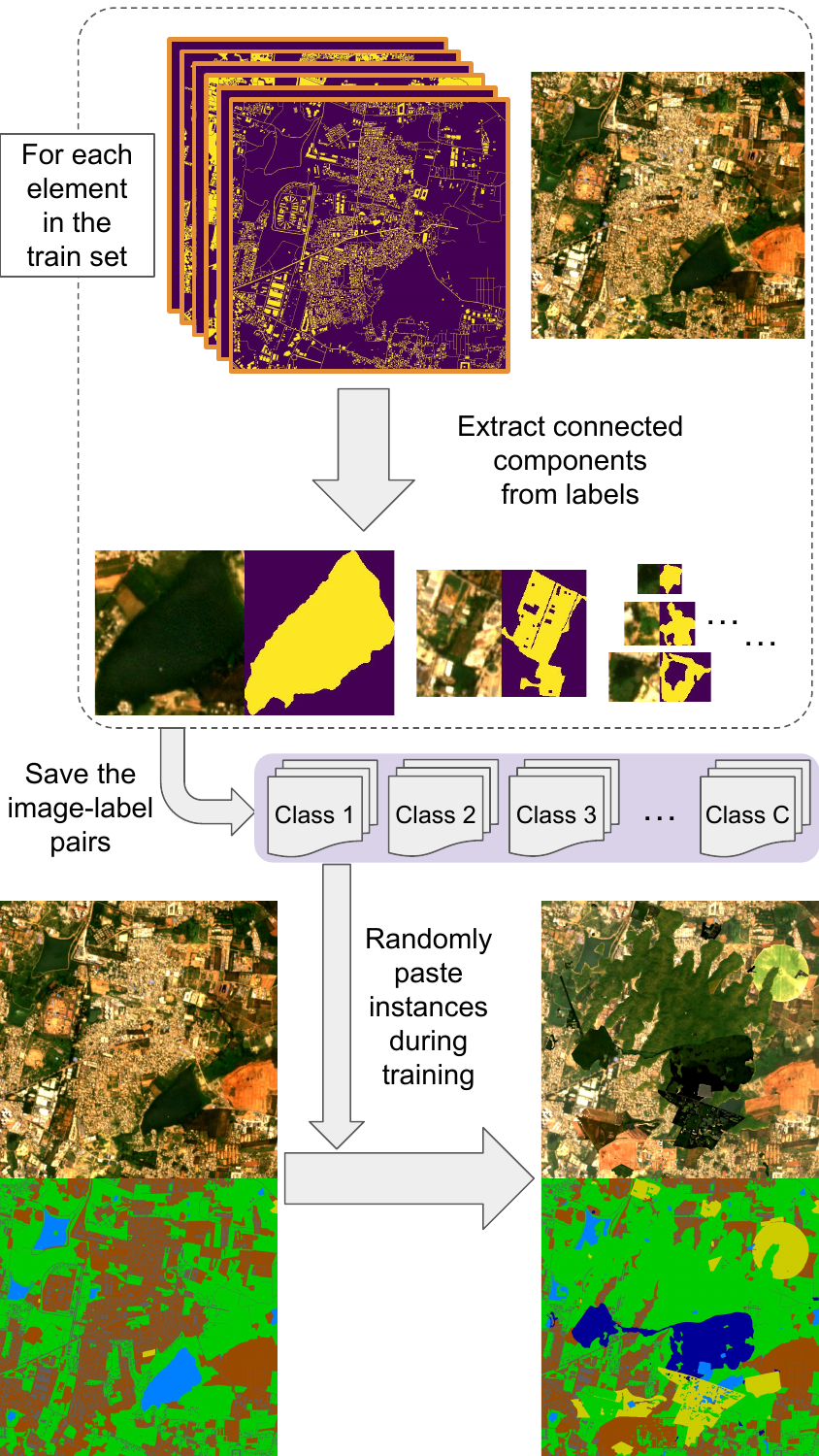}
    \caption{Overview of the Cut-and-Paste augmentation technique. The training set is first processed to extract and save instances that are then randomly pasted during the training phase.}
    \label{fig:cap_overview}
\end{figure}

\section{Methodology}
\label{sec:methodology}
The entire process involves two steps, as shown in Fig.\nobreakspace\ref{fig:cap_overview}. Firstly, we perform an offline preprocessing of the data to extract the instances of interest. Next, we perform an online augmentation of the images by randomly pasting the objects extracted in the previous step.

\subsection{Instance Extraction}
The initial stage is necessary to adapt the Cut-and-Paste augmentation to the problem of semantic segmentation. Unlike instance segmentation, where labels delineate individual objects, semantic segmentation in datasets like DynamicEarthNet~\cite{toker2022dynamicearthnet} provides class information at a pixel level, merged into a single mask image. The key to leveraging the Cut-and-Paste technique in this context lies in the separation of the composite mask into distinct instance masks. This step enables the independent manipulation and pasting of instances in the subsequent stage.

In order to identify individual objects present in an image, we exploit the concept of connected components, which consist of contiguous pixels belonging to the same class. For each image $I$, we separate its corresponding label $L$ into $C$ binary masks $L_c$, one for each class, and apply the connected components algorithm to extract $N_c$ components. In practice, this corresponds to partitioning the label into a set of non-overlapping mask instances ${L}_{c,i}$, each belonging to one of the $C$ classes.

\begin{equation}
    \label{eq:inst_extr}
    L = \bigcup_{c=1}^{C} L_c = \bigcup_{c=1}^{C} \bigcup_{i=1}^{N_c} {L}_{c,i}
\end{equation}

The extracted objects will finally consist of cropped multispectral images and the relative binary masks $(I_{c,i}, L_{c,i})$.
Those are saved in order to create a set of instances containing the extracted objects for each class.

\subsection{Instance Pasting}
The second step is done during training, augmenting each image-label pair on the fly with instances from the set generated in the first stage. We uniformly sample a category $c$ and randomly choose an instance $(I_{c,i}, L_{c,i})$ belonging to that class. The chosen instance is optionally augmented with flips and rotations and then pasted onto the current training sample at a random location. 

We repeat the instance pasting process $N$ times for each training image, where $N$ is a parameter indicating the number of objects to be added. Differently from~\cite{chen2021adversarial}, we allow the instances to overlap, and, as a result, our method can produce unusual scenarios. Although such scenarios may be rare, they are not impossible in reality and can help the trained model to generalize to infrequent real-world cases.

Once the pasting process is concluded, the final image undergoes simple non-destructive transformations, i.e., horizontal/vertical flips and 90-degree rotations.

\section{Experimental Setup}
\subsection{Dataset}
For evaluation, we utilized the Dynamic\-EarthNet dataset~\cite{toker2022dynamicearthnet}, which consists of daily, multispectral satellite images from PlanetLabs. The dataset covers 75 areas of interest around the world, with data spanning over a period of 24 months, from January 2018 to December 2019. Each image includes both RGB and near-infrared bands with a spatial resolution of 3 meters per pixel. The dataset provides pixel-wise monthly semantic segmentation labels of 7 classes: impervious surface, agriculture, forest \& other vegetation, wetlands, soil, water, and snow \& ice. Following the methodology outlined in~\cite{toker2022dynamicearthnet}, we did not consider the snow \& ice class due to its limited representation. Since we focused on fully supervised semantic segmentation, we selected the images that had a corresponding ground truth label, meaning the first day of each month.

The dataset is natively split into training, validation, and test sets. However, as the validation and test sets lack labels, evaluations require submitting predictions to the Codalab platform~\footnote{\href{https://codalab.lisn.upsaclay.fr/competitions/2882}{https://codalab.lisn.upsaclay.fr/competitions/2882}}. To facilitate the validation process, we subdivided the training set into 1200 images for training and 120 images for internal validation, while ensuring that both splits included all the classes and that any area of interest was present in only one of the two splits to prevent data leakage.

\subsection{Evaluation metric}
To evaluate the performance of the models, the mean intersection over union (mIoU) was employed as the primary metric. It is defined as the average of the intersection over union values for the predicted and ground truth labels across all classes:

\begin{equation}
    \label{eq_miou}
    mIoU = \frac{1}{C} \sum_{i=1}^{C} \frac{\mid A_i \cap B_i \mid}{\mid A_i \cup B_i \mid}
\end{equation}

where $C$ is the number of classes, $A_i$ is the ground truth mask for class $i$, and $B_i$ is the prediction mask for class $i$.

\subsection{Training}
Since we are studying the effects of a generally applicable augmentation technique, we limit our comparisons to the use of a basic U-Net model~\cite{ronneberger2015u}, which is widely employed in the literature for semantic segmentation tasks. The U-Net model was trained from scratch on the DynamicEarthNet dataset using an Adam optimizer with a learning rate of 1e-4 for 200 epochs. We saved the model weights at the epoch with the highest mIoU on the internal validation set. To ensure consistency and reproducibility in our experiments, after optimization of the baseline, the hyperparameters remained fixed. We repeated each experiment three times and reported the mean and standard deviation of the mIoU. 

Our baseline model was trained with standard augmentations, including horizontal and vertical flips and random 90-degree rotations, but did not employ the Cut-and-Paste augmentation.

We evaluated the impact of the Cut-and-Paste augmentation in addition to standard augmentations by varying the number of pasted instances per image (N=10, 100, 1000). Furthermore, we explored the influence of augmenting every instance with flips and rotations prior to pasting them onto the image. We refer to those instance augmentations as \textit{\mbox{pre-pasting} augmentations}.

\section{Results}
In this section, we present the results of implementing the Cut-and-Paste augmentation in various configurations. In all the experiments we used Cut-and-Paste in conjunction with the standard augmentations of the baseline to underline its additive contribution. Our results in Tables~\ref{tab:val} and \ref{tab:test} demonstrate notable improvements in model performance over the baseline.

Table\nobreakspace\ref{tab:val} shows the performance of the models on the validation and internal validation sets. When the Cut-and-Paste augmentation was applied with ten instances, the mIoU on the validation set improved, and a slight decrease in performance was seen on the internal validation set.
By increasing the number of pasted instances to 100, the validation improves substantially on both sets, suggesting that the method scales well with an increased number of instances to some extent.
As the number of instances increases to 1000, we observe a drop in mIoU gains, suggesting a point where adding more instances does not yield further improvements and may even be detrimental.

\begin{table}[t]
    \renewcommand{\arraystretch}{1.2}
    \centering
    \caption{Validation performance of the Cut-and-Paste augmentation compared to the baseline. The effects of different numbers of pasted instances and the use of pre-pasting augmentations are considered. The mean and standard deviation are computed over three different runs.}
    \label{tab:val}
    \begin{tabular}{|lccc|}
        \hline
        & Pre-pasting & Val & Internal \\
        & augmentations & mIoU & Val mIoU \\
        \hline
        Baseline & & 34.1 (2.1) & 43.5 (1.0) \\
        \hline
        C\&P N=10 & \cmark & 34.5 (0.8) & 43.4 (0.9) \\
        C\&P N=100 & \cmark & 35.9 (1.1) & 47.4 (0.7) \\
        C\&P N=1000 & \cmark & 35.6 (0.6) & 47.0 (0.6) \\
        \hline
        C\&P N=10 & \xmark & 35.5 (1.3) & 43.2 (0.9) \\
        C\&P N=100 & \xmark & \textbf{36.0} (1.5) &\textbf{47.5} (0.8) \\
        C\&P N=1000 & \xmark & 34.8 (2.0) & 47.0 (0.7) \\
        \hline
    \end{tabular}
\end{table}

\begin{table}[t]
    \renewcommand{\arraystretch}{1.2}
    \centering
    \caption{Test performance of the Cut-and-Paste Augmentation with N=100 pasted instances, with and without pre-pasting augmentations, compared to the baseline.}
    \label{tab:test}
    \begin{tabular}{|lcc|}
        \hline
        & Pre-pasting & Test \\
        & augmentations & mIoU \\
        \hline
        Baseline & & 37.9 \\
        \hline
        C\&P N=100 & \cmark & 42.3 \\
        C\&P N=100 & \xmark & \textbf{44.1} \\
        \hline
    \end{tabular}
\end{table}

The most effective configuration utilized Cut-and-Paste with 100 instances without pre-pasting augmentations, achieving an average highest validation mIoU of 36.0, with a significant improvement over the baseline. The second-best configuration, using the same number of instances but with pre-pasting augmentations, reached a validation mIoU of 35.9. This indicates that pre-pasting augmentations do not necessarily correlate with better performance and that the model can achieve substantial gains with just plain Cut-and-Paste.

\begin{figure}[t]
    \centering
    \captionsetup{justification=centering}
    \begin{subfigure}[t]{0.24\linewidth}
        \includegraphics[width=\linewidth]{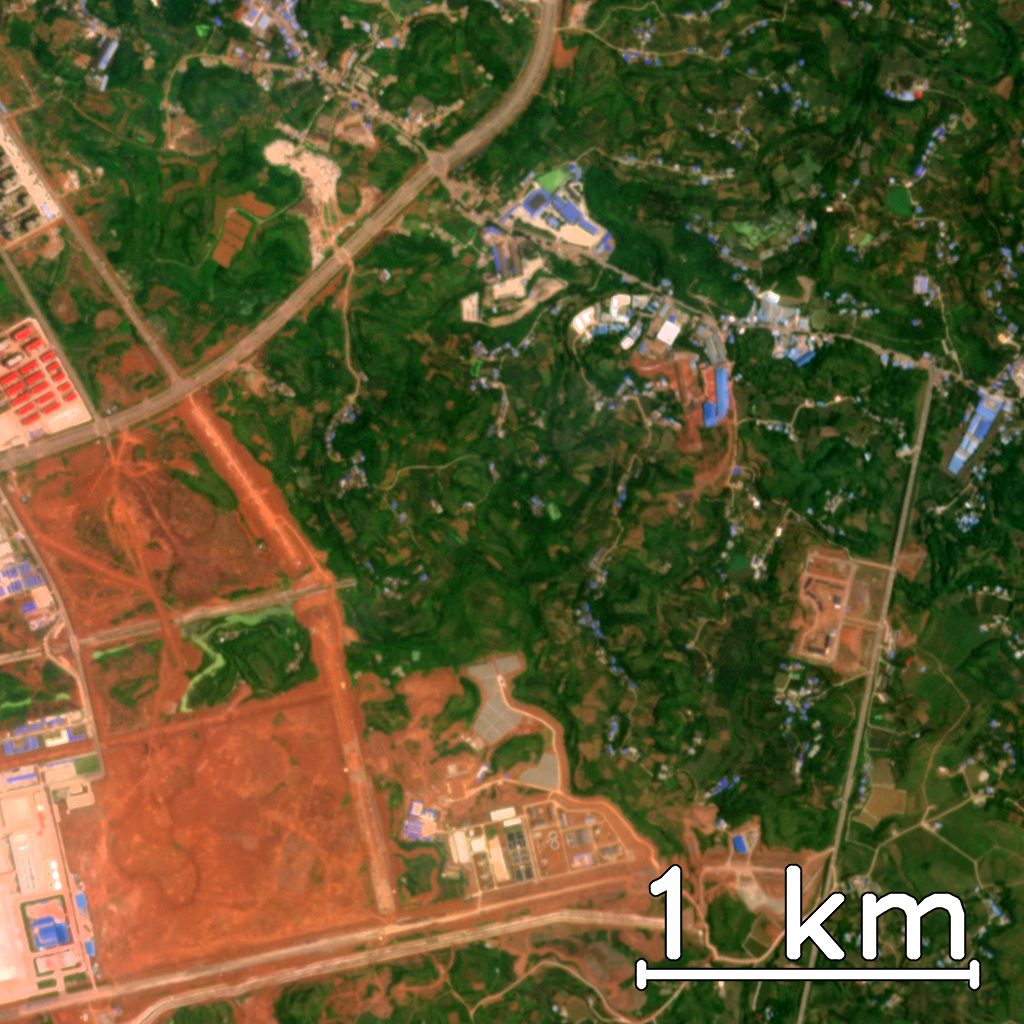}
        \label{fig:img_1}
    \end{subfigure}
    \begin{subfigure}[t]{0.24\linewidth}
        \includegraphics[width=\linewidth]{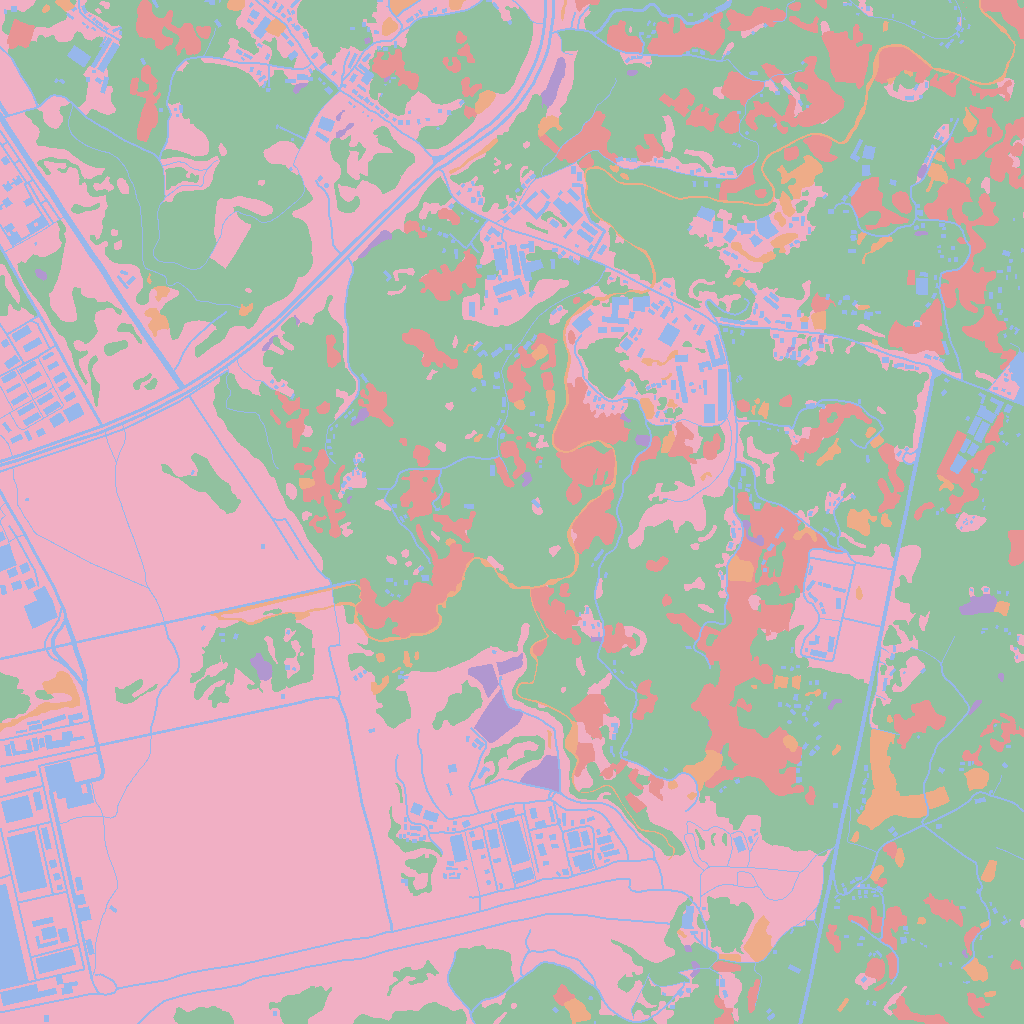}
        \label{fig:gt_1}
    \end{subfigure}
    \begin{subfigure}[t]{0.24\linewidth}
        \includegraphics[width=\linewidth]{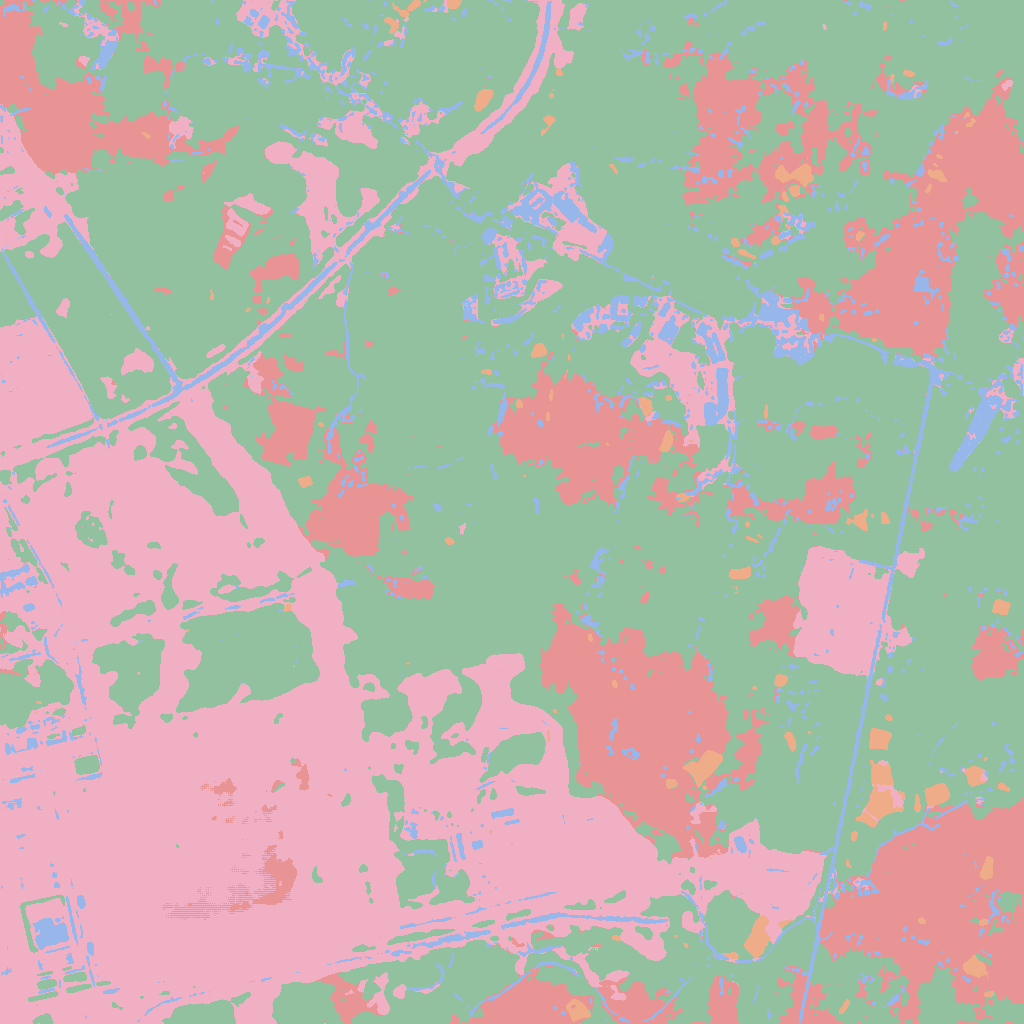}
        \label{fig:baseline_1}
    \end{subfigure}
    \begin{subfigure}[t]{0.24\linewidth}
        \includegraphics[width=\linewidth]{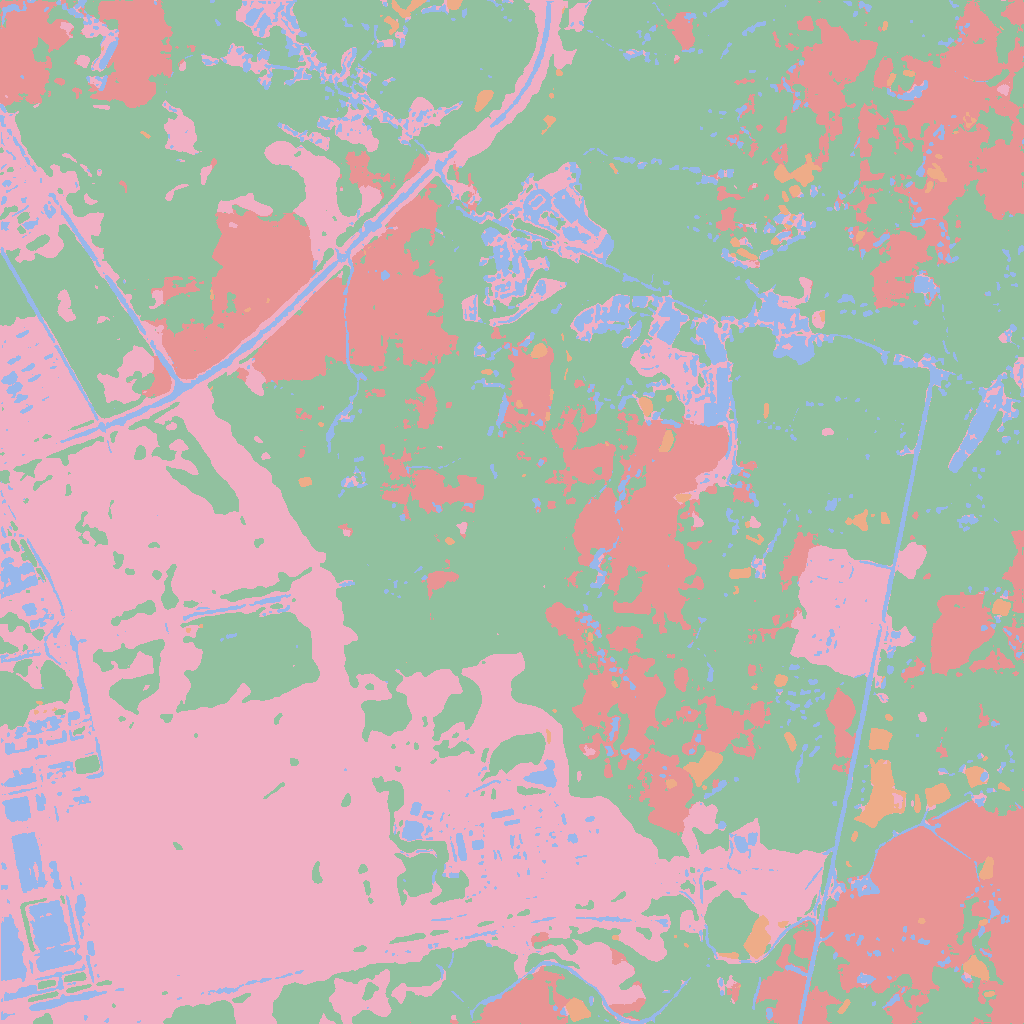}
        \label{fig:cap100_1}
    \end{subfigure}

    \vspace{-10pt}

    \begin{subfigure}[t]{0.24\linewidth}
        \includegraphics[width=\linewidth]{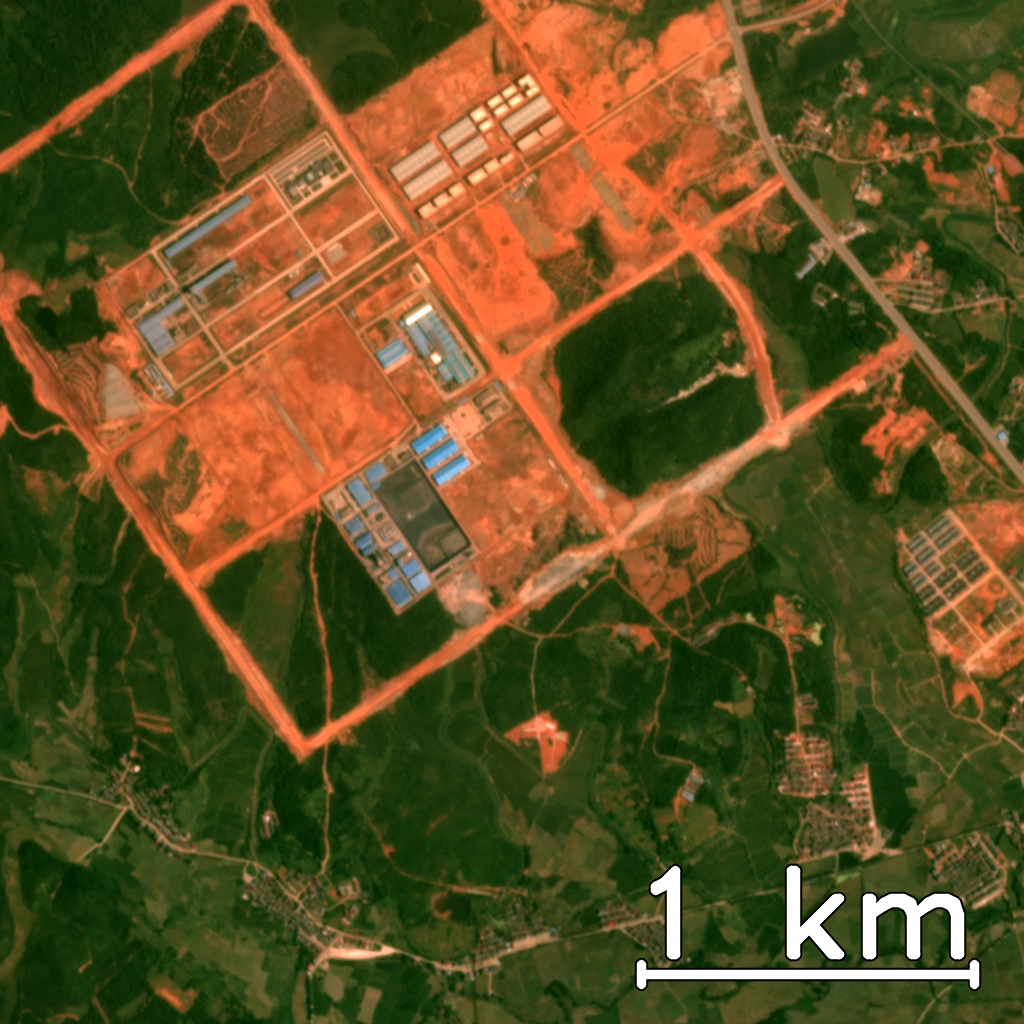}
        \label{fig:img_2}
    \end{subfigure}
    \begin{subfigure}[t]{0.24\linewidth}
        \includegraphics[width=\linewidth]{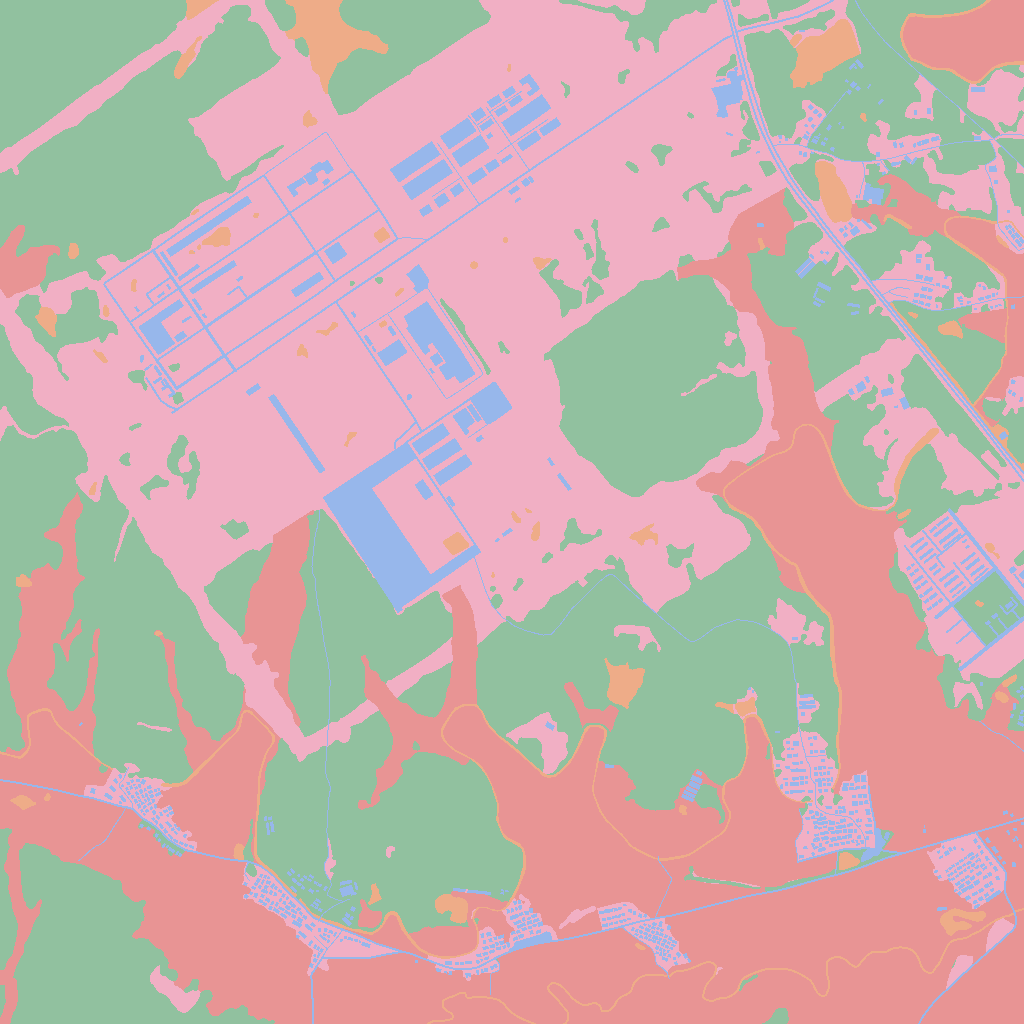}
        \label{fig:gt_2}
    \end{subfigure}
    \begin{subfigure}[t]{0.24\linewidth}
        \includegraphics[width=\linewidth]{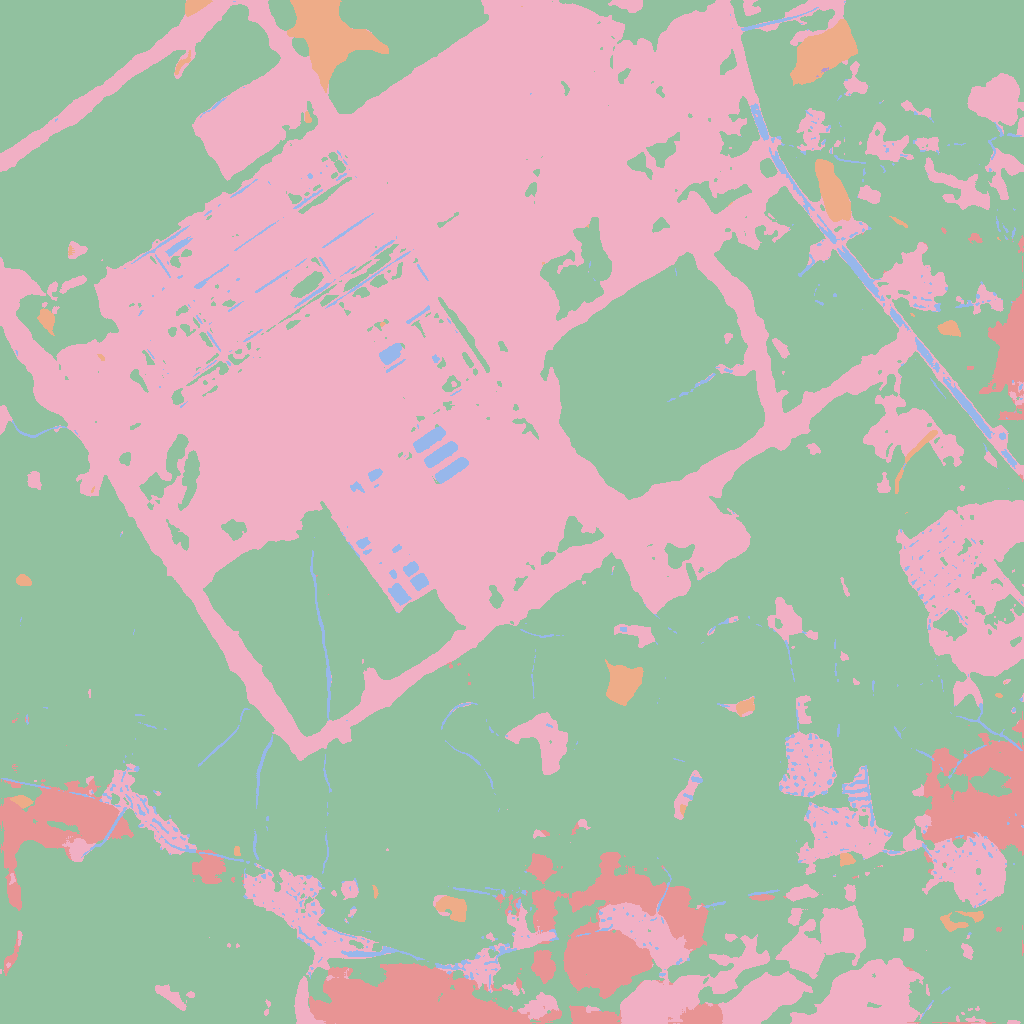}
        \label{fig:baseline_2}
    \end{subfigure}
    \begin{subfigure}[t]{0.24\linewidth}
        \includegraphics[width=\linewidth]{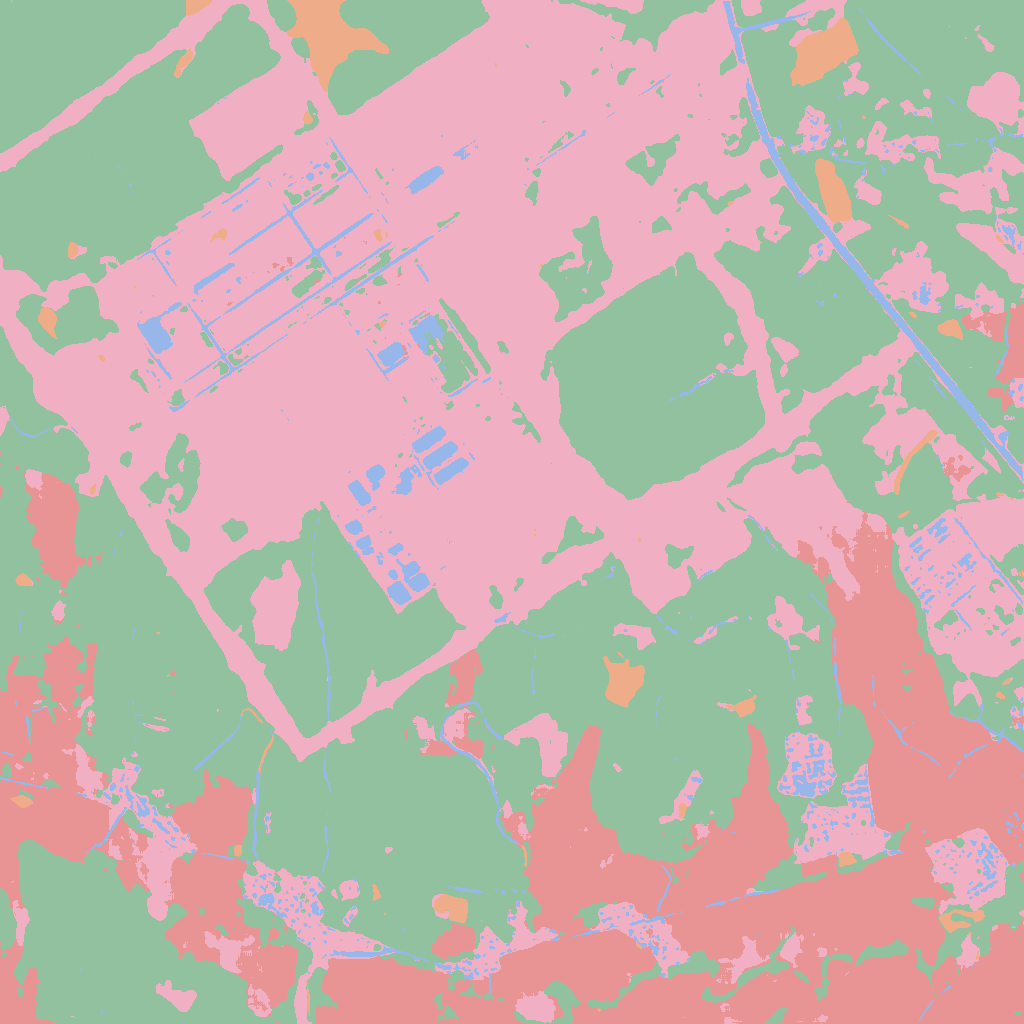}
        \label{cap100_2}
    \end{subfigure}

    \vspace{-10pt}

    \begin{subfigure}[t]{0.24\linewidth}
        \includegraphics[width=\linewidth]{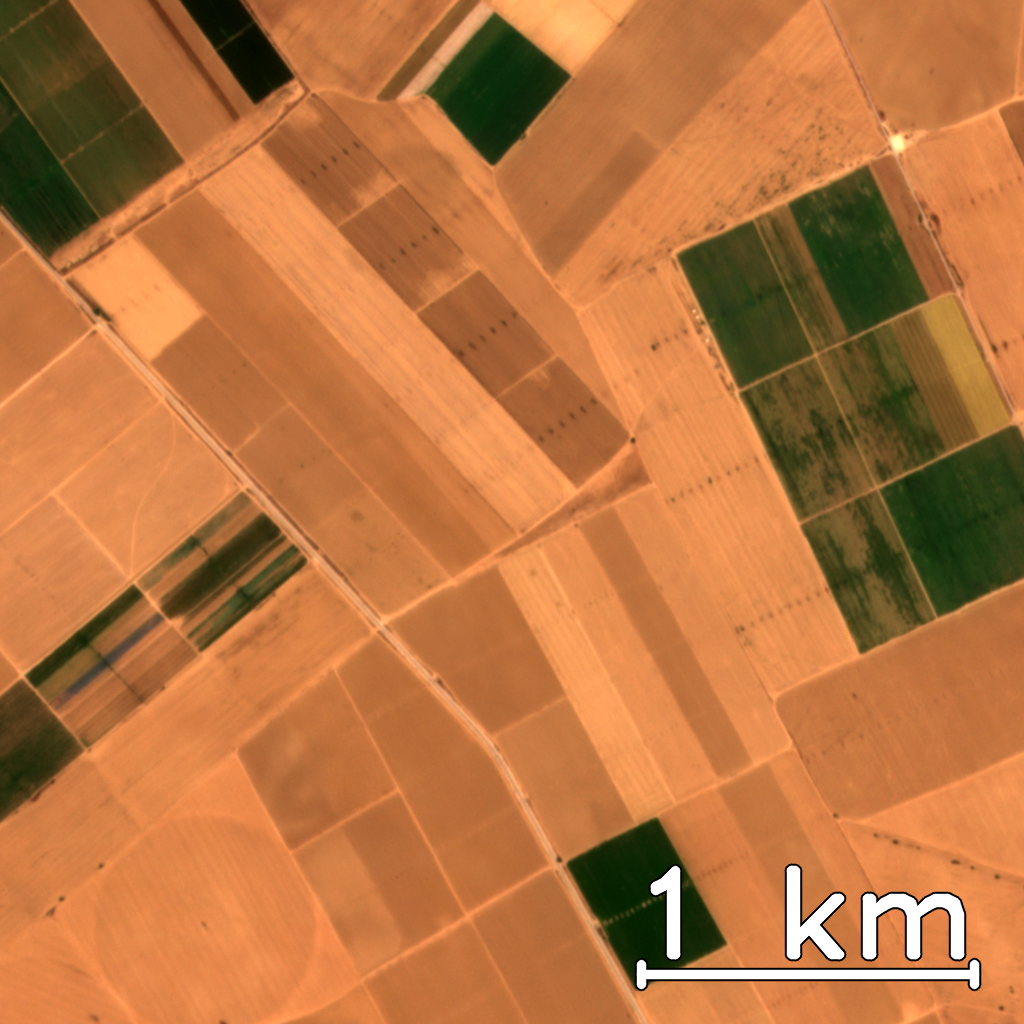}
        \caption{Image}
        \label{fig:img_3}
    \end{subfigure}
    \begin{subfigure}[t]{0.24\linewidth}
        \includegraphics[width=\linewidth]{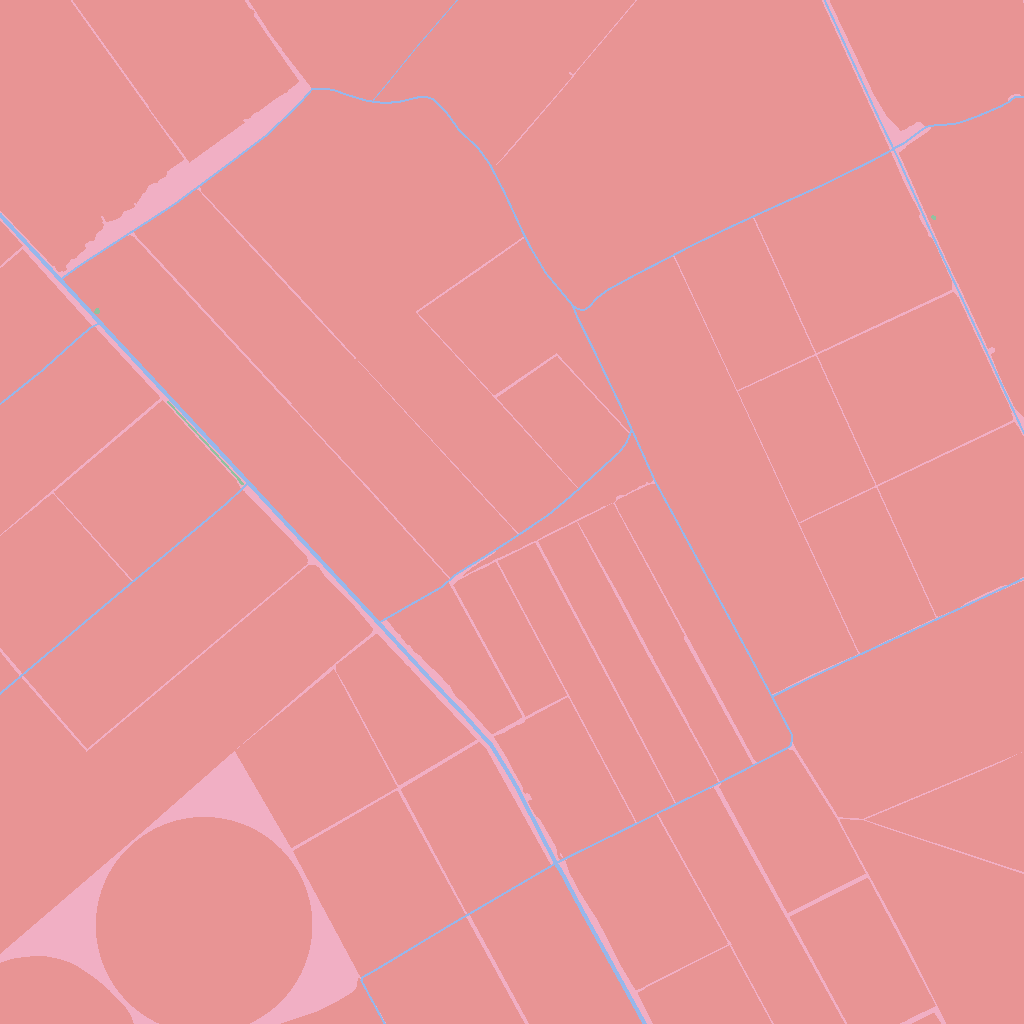}
        \caption{GT}
        \label{fig:gt_3}
    \end{subfigure}
    \begin{subfigure}[t]{0.24\linewidth}
        \includegraphics[width=\linewidth]{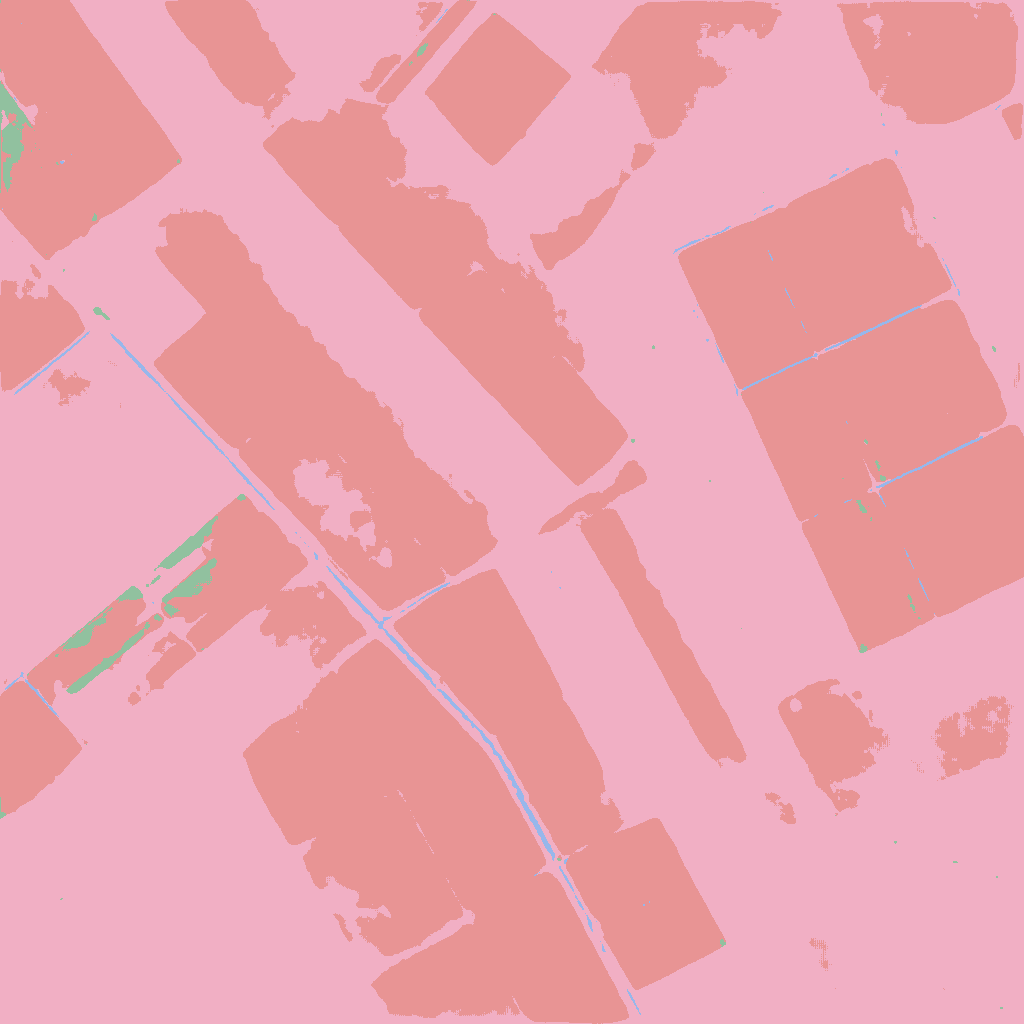}
        \caption{Baseline}
        \label{fig:baseline_3}
    \end{subfigure}
    \begin{subfigure}[t]{0.24\linewidth}
        \includegraphics[width=\linewidth]{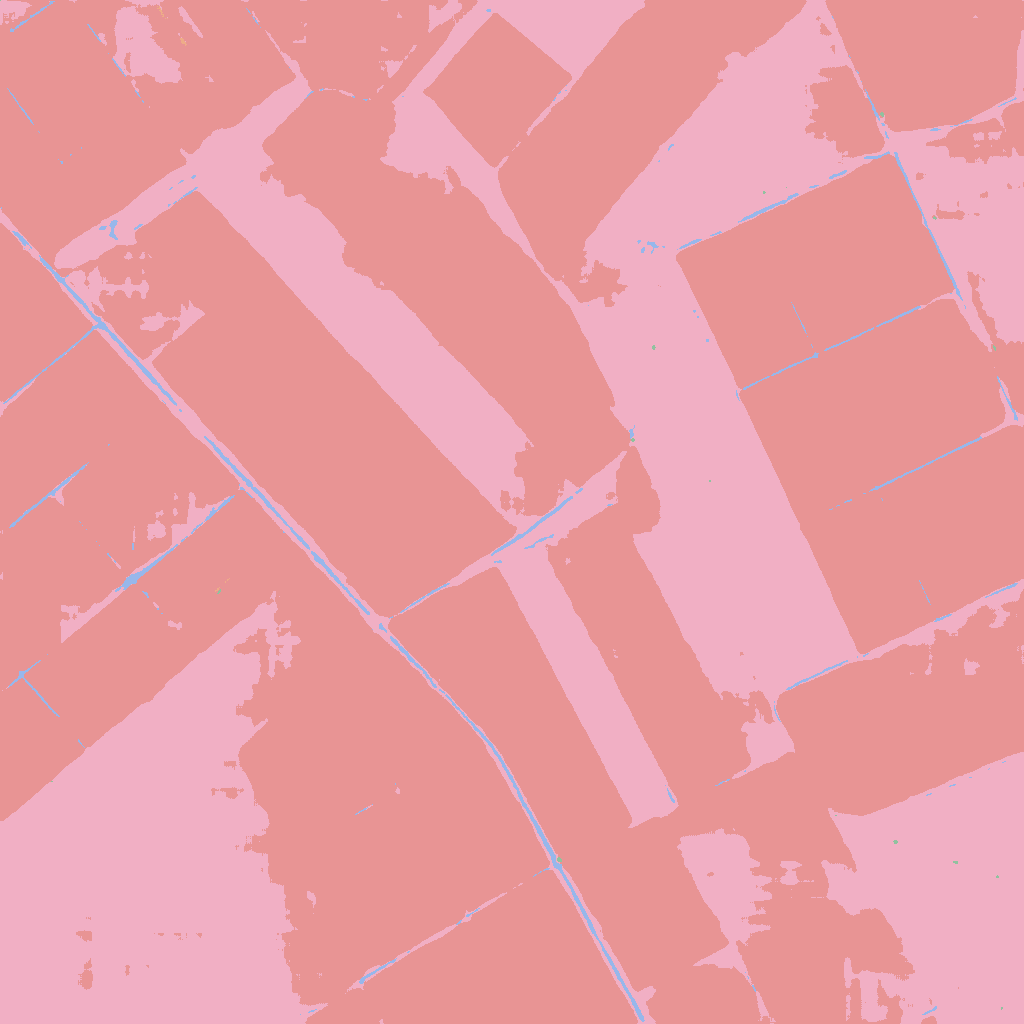}
        \caption{C\&P}
        \label{cap100_3}
    \end{subfigure}

    \begin{subfigure}[t]{\linewidth}
        \includegraphics[width=\linewidth]{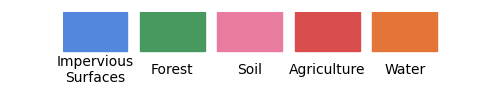}
    \end{subfigure}
    \captionsetup{justification=justified}
    \caption{Qualitative visualization of segmentation results on internal validation images. Subfigure (a) displays the original multispectral satellite image, converted to RGB and enhanced for better visualization, (b) shows the corresponding ground truth segmentation, (c) is the segmentation prediction by the baseline U-Net model, and (d) shows the improved predictions achieved by the U-Net model with Cut-and-Paste augmentation (N=100 without pre-pasting augmentations). 
    }
    \label{fig:main}
\end{figure}

Fig.\nobreakspace\ref{fig:main} provides visual comparisons of the segmentation results between the baseline and the best-performing model on two images from the internal validation set. The improvements are particularly notable in the impervious surface and agriculture classes, where our augmentation approach significantly enhances the model's predictive accuracy.

For the test set evaluation (Table\nobreakspace\ref{tab:test}), only the two best configurations were further assessed. Here, the performance gains were more pronounced: the configuration with 100 pasted instances without pre-pasting augmentations reached a test mIoU of 44.1, and the one with pre-pasting augmentations achieved 42.3, both significantly surpassing the baseline's mIoU of 37.9. 
These results highlight the general efficacy of the Cut-and-Paste augmentation technique and suggest that the number of instances and the use of pre-pasting augmentations need to be carefully calibrated to obtain the greatest improvement.

\section{Conclusion}

In this study, we have explored the application of a Cut-and-Paste augmentation technique for semantic segmentation of satellite imagery. We have adapted this technique, which usually requires labeled instances, to the case of semantic segmentation by leveraging the connected components in the semantic labels. We have demonstrated that this augmentation enhances the data diversity and variability, improving the generalization capabilities of the segmentation model. Our experiments, performed using a simple U-Net model, have shown that the Cut-and-Paste augmentation provides a significant performance increase, bringing the mIoU score from 37.9 to 44.1 on the DynamicEarthNet test set. Our approach offers a simple and effective solution for generating new semantic segmentation data of satellite images without requiring additional manual annotations. Future work could further refine this technique and explore its applicability to other remote sensing tasks, such as Change Detection.

\bibliographystyle{IEEEbib}
\bibliography{refs}

\end{document}